\documentclass[paperpaper, 10 pt, conference]{ieeeconf}  

\IEEEoverridecommandlockouts                              

\usepackage{graphics} 
\usepackage{subfigure}
\usepackage{epsfig} 
\usepackage{mathptmx} 
\usepackage{times} 
\usepackage{amsmath} 
\usepackage{amssymb}  
\usepackage{booktabs}	%
\usepackage{balance}
\usepackage{xcolor} 
\usepackage{cite}
\usepackage{mathtools}

\title{\LARGE \bf
Model Predictive Spherical Image-Based Visual Servoing On $SO(3)$ for Aggressive Aerial Tracking
}

\author{Chao Qin$^{1}$, Qiuyu Yu$^{2}$, and Hugh H.T. Liu$^{1}$
\thanks{$^{1}$Chao Qin and Hugh H.T. Liu are with the University of Toronto Institute for Aerospace Studies, Toronto, Canada
{\tt\small chao.qin@mail.utoronto.ca, hugh.liu@utoronto.ca}}
\thanks{$^{2}$Qiuyu Yu is with the School of Astronautics and Aeronautics, Shanghai Jiao Tong University, Shanghai, China
	{\tt\small joeyyu@sjtu.edu.cn}}
}

\begin{document}

\maketitle
\thispagestyle{empty}
\pagestyle{empty}

\begin{abstract}
This paper presents an image-based visual servo control (IBVS) method for a first-person-view (FPV) quadrotor to conduct aggressive aerial tracking. There are three major challenges to maneuvering an underactuated vehicle using IBVS: (i) finding a visual feature representation that is robust to large rotations and is suited to be an optimization variable; (ii) keeping the target visible without sacrificing the robot's agility; and (iii) compensating for the rotational effects in the detected features. We propose a complete design framework to address these problems. First, we employ a rotation on $SO(3)$ to represent a spherical image feature on $S^{2}$ to gain singularity-free and second-order differentiable properties. To ensure target visibility, we formulate the IBVS as a nonlinear model predictive control (NMPC) problem with three constraints taken into account: the robot's physical limits, target visibility, and time-to-collision (TTC). Furthermore, we propose a novel attitude-compensation scheme to enable formulating the visibility constraint in the actual image plane instead of a virtual fix-orientation image plane. It guarantees that the visibility constraint is valid under large rotations. Extensive experimental results show that our method can track a fast-moving target stably and aggressively without the aid of a localization system.

\begin{keywords}
Visual servoing, aerial systems: mechanics and control, model predictive control.
\end{keywords}

\end{abstract}

\section{INTRODUCTION}

While human pilots employ visual information from a front-looking camera of a drone to accomplish complicated missions such as racing and chasing \cite{pfeiffer2021human}, most vision-based unmanned aerial vehicles (UAVs) require an intermediate localization module to transform image data to position feedback \cite{ji2022elastic, wang2021visibility}. Such transformation not only increases time delay but also makes the system vulnerable to image blurs, poor lighting conditions, and modeling errors \cite{potena2017effective}. Moreover, it will take up large amounts of computational resources that may not be affordable for small UAVs. Therefore, it is of great importance to investigate image-based visual servo control (IBVS) that can directly exploit visual information for robot control \cite{corke2010spherical}.

\begin{figure}[!t]
	\centering\
	\includegraphics[width=0.4\textwidth]{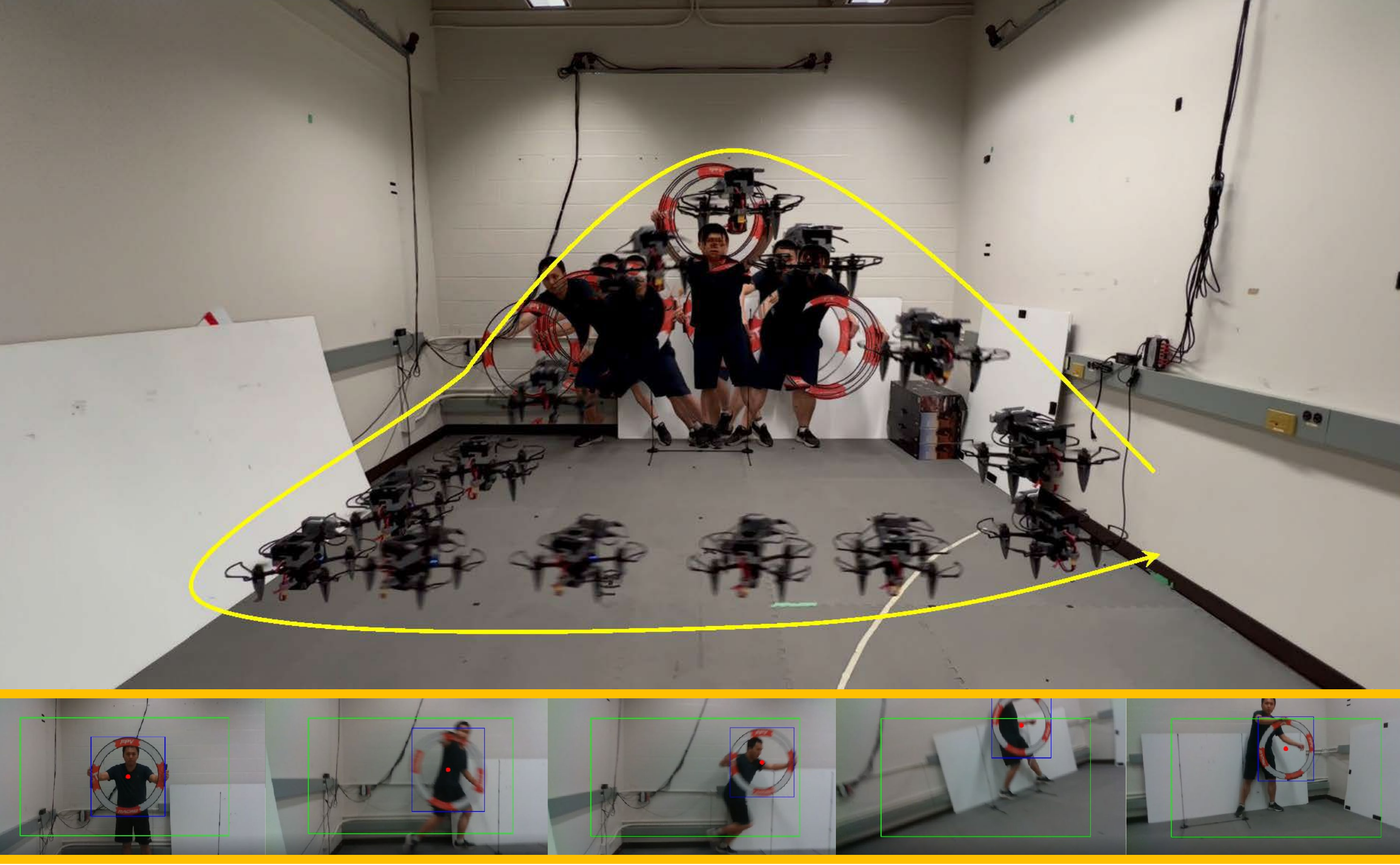}
	\caption{Our IBVS method can track a fast-moving target and keep it visible during large rotations. We can ensure that the produced aggressive maneuvers will not jeopardize the target visibility. In this experiment, the circular racing gate is detected by a deep-learning-based object detector (blue box), and its center is kept inside the specified image bound (green box) at all times.}
	\label{fig_moving_gate}
\end{figure}

However, IBVS exhibits a degraded performance on a quadrotor since the 6-DOF commands from a classic IBVS method \cite{chaumette2006visual} cannot be perfectly executed by an underactuated platform with only 4 DOF \cite{li2021image}. Furthermore, since the camera is rigidly connected to the vehicle, the coupling of the quadrotor's translational and rotational motions can significantly hinder the target visibility. To mitigate these effects, previous methods operate the quadrotor in a near-hover state \cite{mcfadyen2013aircraft, zhang2021robust, li2021image}, which seriously restricts its agility, or implicitly assume that controlling the heading direction suffices to keep the target visible \cite{guo2020image}, which is impractical under large pitching motions. This paper does not rely on any assumption regarding target visibility, because it can be satisfied by fully considering the impact of the quadrotor dynamics on the motion of the image feature. As shown in Fig. \ref{fig_moving_gate}, the target can be kept inside the camera field of view (FOV) under aggressive maneuvers. 

\begin{figure}[!t]
	\centering\
	\includegraphics[width=0.35\textwidth]{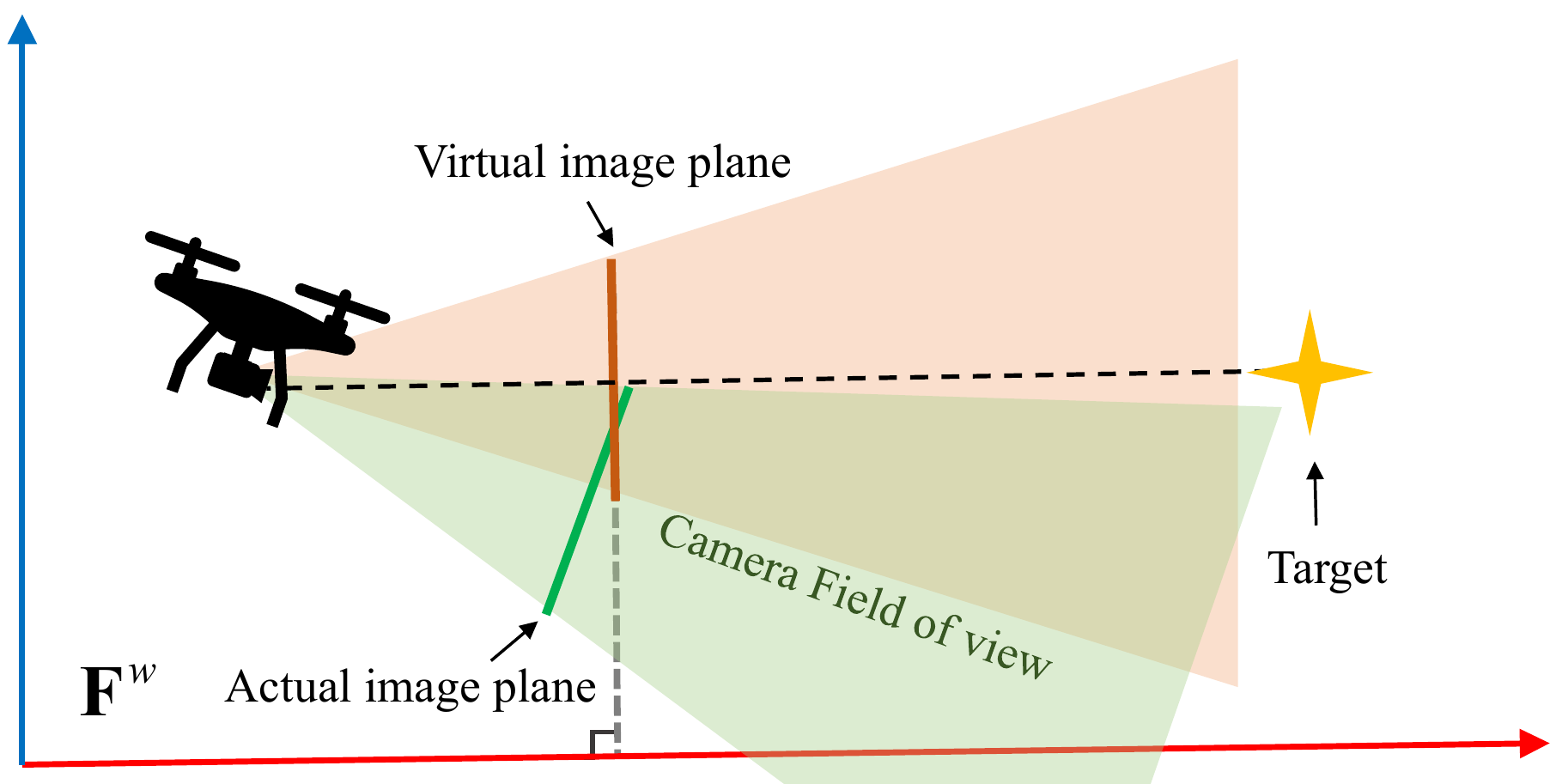}
	\caption{The underlying problem of formulating visibility constraints in a virtual image plane instead of the actual image plane. In this example, even though the target has already left the camera FOV, the visibility constraint defined in the virtual image plane can still be satisfied, meaning that it is already an invalid constraint.}
	\label{fig_fov}
\end{figure}

Recently, formulating IBVS on a virtual fixed-orientation image plane has been validated in autonomous landing tasks\cite{mcfadyen2013aircraft, sheng2019image, zhang2021robust}. However, these methods will fail in agile flights since their visibility constraints will become useless when the camera rotation is large. An example is illustrated in Fig. \ref{fig_fov}. Additionally, they demand an extra position \cite{potena2017effective}, velocity \cite{roque2020fast}, or attitude \cite{mcfadyen2013aircraft, zhang2021robust} controller which has no guarantee to fulfill the perceptive constraints. This paper solves the invalid visibility constraint problem and removes the need for a position/velocity/attitude controller. 

In this paper, we present a spherical IBVS method based on nonlinear model predictive control (NMPC) to enable aggressive aerial tracking with a quadrotor equipped with a front-looking camera and an inertial measurement unit (IMU). To tackle large rotations, we project the 2D image feature to the unit sphere and utilize a rotation as the underlying representation of the resulting 3D unit vector. This feature representation not only enjoys the internal passivity-like property that mitigates the impact of underactuation \cite{hamel2002visual} but also addresses the non-smooth vector field problem as stated by the ``hairy ball theorem'' \cite{lavalle2006planning} when optimizing a state variable (3D unit vector) on the $S^{2}$ manifold. Then, an optimal control problem is constructed which takes the image kinematics and quadrotor dynamics as equality constraints and takes the actuation limits and target visibility as inequality constraints. Additionally, we propose a time-to-collision (TTC) constraint to reduce overshoots caused by high tracking speeds. 

To sum up, the novel contributions of this paper are:
\begin{itemize}
	\item {We present a robust IBVS algorithm for agile underactuated robots. The target visibility can be guaranteed without any assumption on the robot motion. Our work will be available online \footnote{https://github.com/ChaoqinRobotics}}
	\item {We introduce a rotation-based feature representation that is suited for model predictive IBVS, a.k.a. visual predictive control (VPC).}
	\item {We propose a novel attitude-compensation scheme to enable formulating the visibility constraint in the actual image plane.}
\end{itemize}

\section{Related Works}
Aerial tracking algorithms can be divided into position-based methods and image-based methods. In this section, we restrict most of our attention to image-based methods.

\subsection{Position-based Methods}
Position-based methods assume that the locations of the robot and target are accessible via GPS \cite{zhang2010vision} or a vision-based state estimator \cite{ji2022elastic}. Given these measures, they solve the tracking problem in two steps: trajectory planning and position tracking. To improve flight safety, many cost functions and constraints are raised which include target visibility \cite{wang2021visibility}, collision avoidance \cite{han2021fast}, and distance keeping \cite{ji2022elastic}. Position-based visual servo control (PBVS) is a special class of position-based methods which utilize relative pose estimate between the robot and the target for target tracking \cite{popova2016position}. Recently, trajectory planning has also been widely applied in PBVS to boost the tracking performance \cite{sheckells2016optimal, potena2017effective}. However, motion planning and control in the Cartesian space requires accurate model parameters of the camera and robot \cite{guo2020image}. Furthermore, it is challenging to obtain high-accuracy position estimate from low-quality images.

\subsection{Image-based Methods}
Fortunately, we are still able to identify some objects or special patterns from low-quality images. These visual features can be used in image-based methods to continue the flight mission, and strong robustness to camera calibration errors and distance errors can be achieved \cite{corke2010spherical}. However, features captured by an onboard camera contain both spatial information about the objects and the robot's attitude, and one needs to decouple these two effects to achieve global convergence \cite{guo2020image}. Image-based methods can be classified into two categories based on different camera-rotation decoupling schemes: invariant feature approaches and virtual camera approaches.

%

Invariant feature approaches are based on the observation that
rotational motions will not change the shape of an object projected on the unit sphere \cite{tahri2013robust}. Consequently, if the shape information is modeled as a feature, its kinematics will be invariant to the camera's rotation. Tahri et al. \cite{tahri2013robust} provided a systematic analysis of a control law using invariant features. Fomena et al. \cite{fomena2011distance} utilized the Cartesian distance between the spherical projections of three points to design a controller that is robust to point-range errors. Guo et al. \cite{guo2020image} showed that the invariant features together with the heading angle can serve as the flat outputs of the differentially-flat vision-based quadcopter system. Based on this property, a geometric tracking controller can be developed to accomplish a multiple-openings traversing task. However, the visibility constraint is not considered in these methods since it is difficult to map invariant features to points in the image plane.

Virtual camera approaches compensate for the camera's rotation using the roll and pitch estimates from an IMU. Therefore, features defined in the virtual image plane can be rotation-invariant, so as their kinematics  \cite{zhang2021robust, sheng2019image, li2021image}. Li et al. \cite{li2021image} present an adaptive backstepping controller to regulate the quadrotor's translation and heading relative to a planar target with arbitrary orientation. Zhang et al. \cite{zhang2021robust} took image moments defined in a level virtual image plane as state variables for NMPC. Similarly, Mcfadyen et al. \cite{mcfadyen2013aircraft} constructed an optimal control problem on a virtual spherical camera model with a single-point feature. The major advantage of virtual camera approaches is the convenience of formulating the visibility constraint. However, previous methods define the visibility constraint in the virtual image plane which fails to prevent the feature to escape the camera FOV in large rotations as shown in Fig. \ref{fig_fov}. To make visibility constraint effective in aggressive maneuvers, we propose to keep the detected feature in the actual image plane and manage attitude compensation in the objective function.

\subsection{Feature Representation and Parametrization}
Classic IBVS methods represent features in the homogenous coordinate \cite{chaumette2006visual}. Although this representation is simple and intuitive, it cannot generate the optimal Cartesian path for large rotation around the optical axis \cite{liu2020robust}, which results in the well-known camera retreat problem \cite{corke2011robotics}. An effective solution is to transform the feature in different coordinate systems. Chaumette et al. \cite{chaumette2007visual} expressed features in the polar coordinate to obtain stable performance in large rotations. Projecting features to the spherical coordinate can also handle the camera retreat problem \cite{corke2011robotics}. Hamel et al. \cite{hamel2002visual} proved that this projection is amendable for underactuated robots since it preserves the passivity-like property of the image kinematics. As a result, spherical IBVS has become one of the most popular controllers for quadrotors \cite{mebarki2015nonlinear, mcfadyen2013aircraft}.

In spherical IBVS, there are multiple ways to parameterize a point on the spherical surface, but not all of them are suited to be an optimization variable. For example, Corke \cite{corke2010spherical} and Mcfadyen et al. \cite{mcfadyen2013aircraft} used 2D-angles which corresponds to the minimal representation. However, this parametrization induces a singularity in the image kinematics around certain angles. Hamel et al. \cite{hamel2002visual} and Mebarki et al. \cite{mebarki2015nonlinear} applied a 3D unit vector to over-parameterize features on the unit sphere. The underlying problem of this parametrization is the non-smooth vector fields on $S^{2}$, a consequence of the ``hairy ball theorem'' \cite{lavalle2006planning}. Since this property is not desired for optimization solvers, Bloesch et al. \cite{bloesch2015robust} proposed to use a rotation to parameterize a 3D unit vector to get a smooth vector field, and they validated this methodology in vision-based pose estimation. In this paper, we introduce this feature parametrization \cite{bloesch2015robust} to NMPC-based spherical IBVS to obtain stable optimization performance.

\section{Spherical Image Kinematics On $SO(3)$}
In this section, we introduce spherical imaging and the rotation-based feature parametrization.

\begin{figure}[!t]
	\centering
	\includegraphics[width=2.0in]{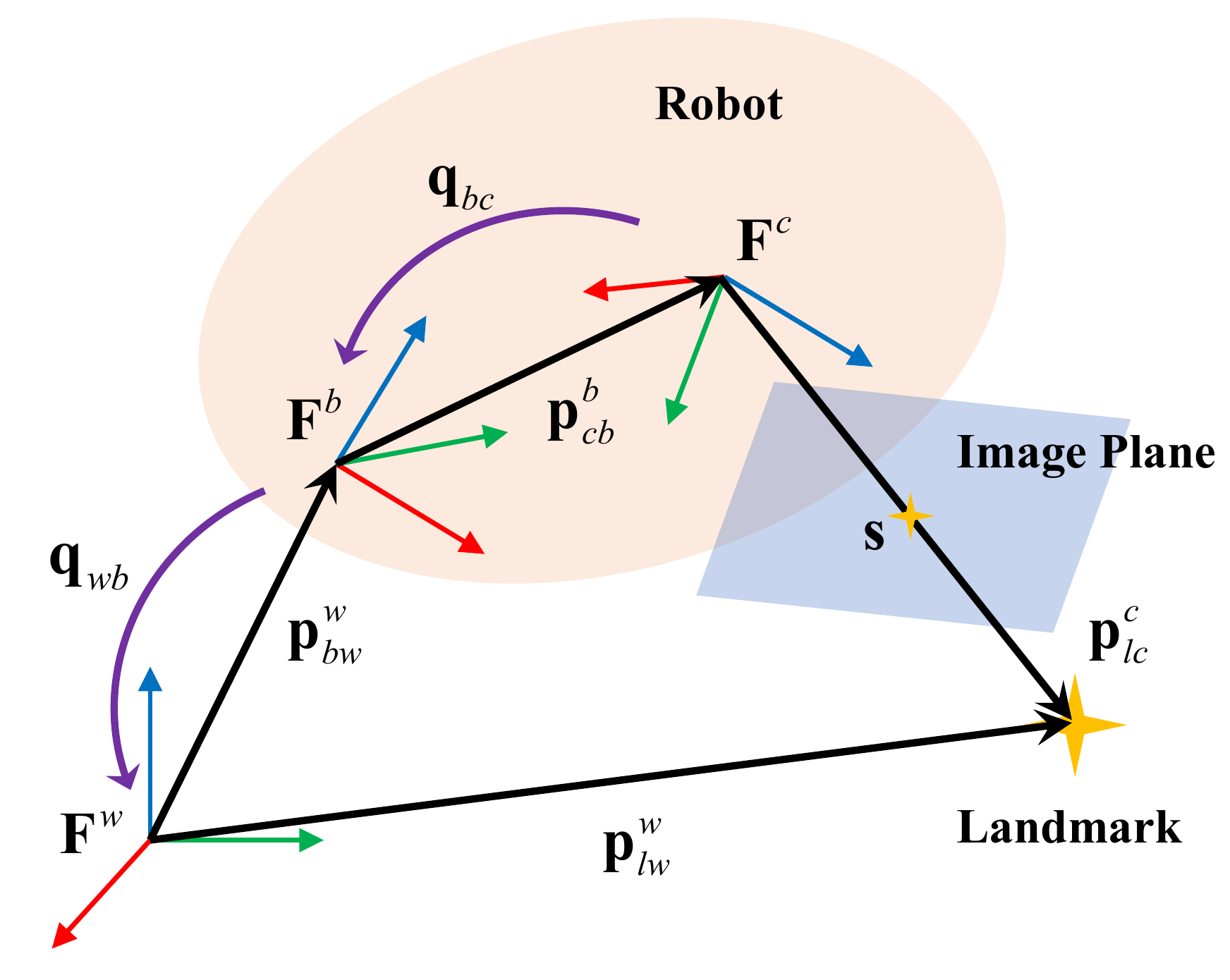}
	\caption{A schematics representing the world frame $\mathcal{F}^{w}$, body frame $\mathcal{F}^{b}$, and camera frame $\mathcal{F}^{c}$. The standard basis is colored as \{\textcolor{red}{$\mathbf{x}$}, \textcolor{green}{$\mathbf{y}$}, \textcolor{blue}{$\mathbf{z}$}\}. The relative pose from the camera frame to the body frame is expressed as $(\mathbf{p}^{b}_{cb},\mathbf{q}_{bc})$. Similarly, $(\mathbf{p}^{w}_{bw},\mathbf{q}_{wb})$ denotes the pose from the body frame to the world frame. A landmark point $l$ located at $\mathbf{p}^{w}_{lw}$ is projected onto image coordinate at $\mathbf{s}$.}
	\label{fig_frame}
\end{figure}

\subsection{Notation}
As shown in Fig. \ref{fig_frame}, three coordinate frames are used throughout the paper: the world frame, $\mathcal{F}^{w}$, the body frame $\mathcal{F}^{b}$, and the camera frame, $\mathcal{F}^{c}$. Let $\mathbf{p}^{w}_{lw}\in \mathbb{R}^3$ be the global position of a landmark $l$, which reads ``the vector from the origin of frame $\mathcal{F}^{w}$ to point $l$, expressed in frame $\mathcal{F}^{w}$''. A rotation is expressed as a quaternion $\mathbf{q}\in SO(3)$ with corresponding rotation matrix denoted as $\mathbf{C}(\mathbf{q})\in \mathbb{R}^{3\times3}$. We use both $\mathbf{q}_{wc}$ and $\mathbf{C}(\mathbf{q}_{wc})$ to represent the rotation from frame $\mathcal{F}^{c}$ to frame $\mathcal{F}^{w}$. An operator $\otimes$ is used to denote the Hamilton product of two quaternions. A point in the image plane expressed in the homogeneous coordinate is denoted by $\mathbf{s}=(u,v,1)$. The skew-symmetric matrix of a 3D vector is written by $(\cdot)^{\times}$.

\begin{figure}[!t]
	\centering
	\includegraphics[width=2.5in]{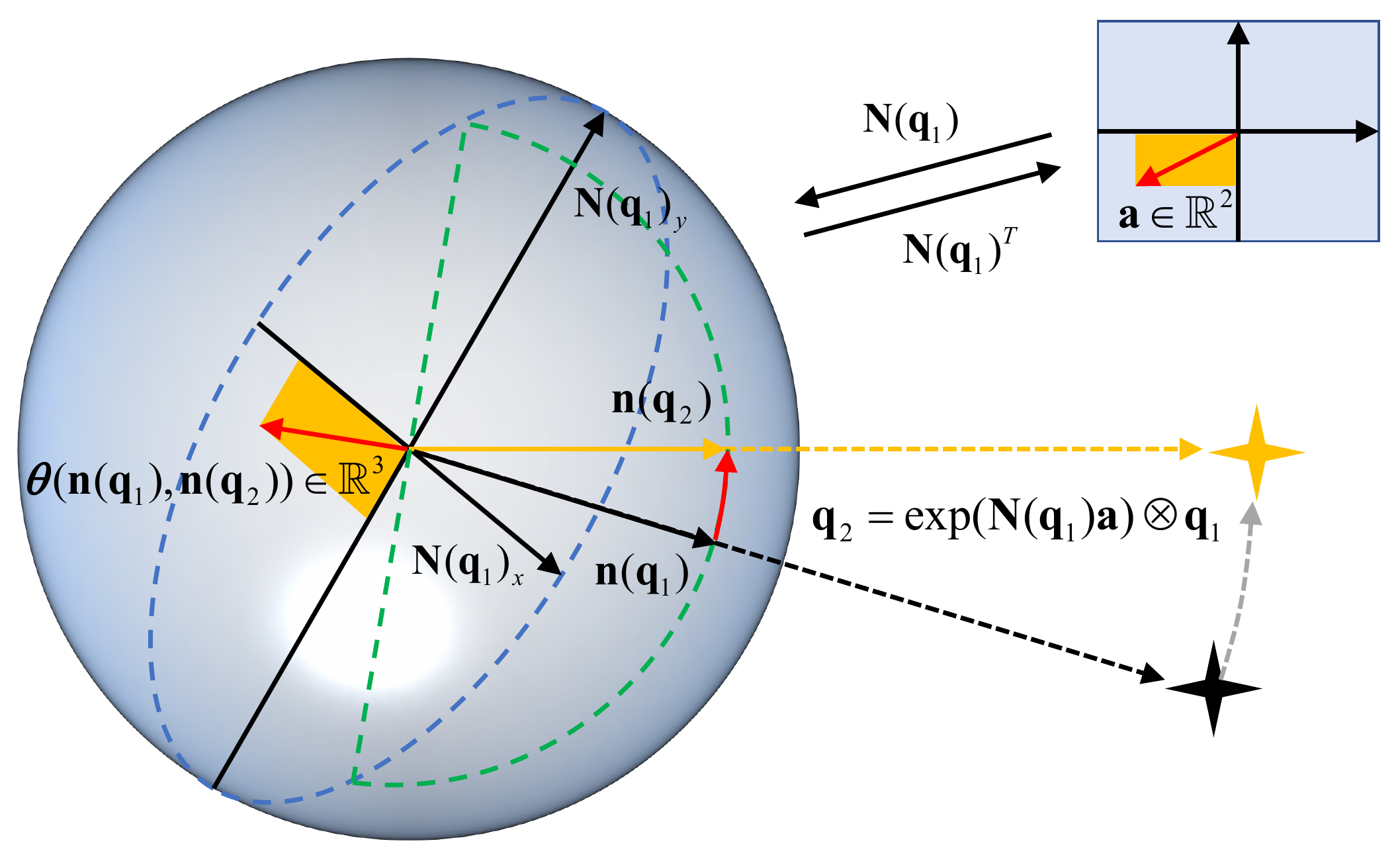}
	\caption{Representation of a point on the spherical coordinate following \cite{bloesch2015robust}. A 3D unit vector $\boldsymbol{\rho}_1$ on the unit sphere pointing toward to the landmark is transformed into an equivalent form $\mathbf{n}(\mathbf{q}_1)$ where $\mathbf{q}_1$ is its underlying rotation representation. The matrix $\mathbf{N}(\mathbf{q}_1)=[\mathbf{N}(\mathbf{q}_1)_x, \mathbf{N}(\mathbf{q}_1)_y]$ determines the orthonormal vector that spans the tangent space such that operators $\boxplus$ and $\boxminus$ can be uniquely defined on the manifold. The operator $\boxminus$ inputs two 3D unit vectors and outputs their difference $\mathbf{a}\in\mathbb{R}^{2}$. Conversely, the operator $\boxplus$ takes an element from $\mathbb{R}^{2}$ and rotates $\mathbf{q}_1$ to $\mathbf{q}_2$. The new location of the point in the unit sphere can be obtained by $\boldsymbol{\rho}_2=\mathbf{n}(\mathbf{q}_2)$.}
	\label{fig_sphere}
\end{figure}

\subsection{Spherical Imaging}
Spherical imaging applies a unit sphere as the image ``plane'' \cite{corke2010spherical}. Assume that the intrinsic parameter matrix \cite{barfoot2017state}, $\mathbf{K}\in \mathbb{R}^{3\times3}$, of the pinhole camera is known. A detected feature $\mathbf{s}$ can be projected on the unit sphere by using $\boldsymbol{\rho} = \mathbf{v}/||\mathbf{v}||$ where $\mathbf{v} = \mathbf{K}^{-1} \mathbf{s}\in\mathbb{R}^{3}$.


\subsection{Rotation-Based Feature Representation}
A rotation $\mathbf{q}\in SO(3)$ is utilized as the underlying representation of a 3D unit vector $\boldsymbol{\rho}\in S^{2}$. We first define the following quantities:
\begin{align}
	\mathbf{n}(\mathbf{q}) & =\mathbf{C}(\mathbf{q})\mathbf{e}_{z},\\
	\mathbf{N}(\mathbf{q}) & =[\mathbf{C}(\mathbf{q})\mathbf{e}_{x},\mathbf{C}(\mathbf{q})\mathbf{e}_{y}],
\end{align}
where $\mathbf{I}=[\mathbf{e}_{x}, \mathbf{e}_{y},\mathbf{e}_{z}]\in\mathbb{R}^{3\times 3}$. We can compute the 3D unit vector $\boldsymbol{\rho}$ from $\mathbf{q}$ by rotating $\mathbf{e}_{z}$ by $\mathbf{q}$, i.e., $\boldsymbol{\rho}=\mathbf{n}(\mathbf{q})$. The matrix $\mathbf{N}(\mathbf{q})$ contains the rotated $\mathbf{e}_{x}$ and $\mathbf{e}_{y}$ vectors that spans the tangent space around $\boldsymbol{\rho}$ as shown in Fig. \ref{fig_sphere}. Given this tangent space, we are able to define the following operators:
\begin{align}
\boxplus:&SO(3)\times\mathbb{R}^{2}\rightarrow SO(3), \\
&\mathbf{q},\mathbf{a}\mapsto\exp(\mathbf{N}(\mathbf{q})\mathbf{a})\otimes\mathbf{q}, \\ 
\boxminus:&SO(3)\times SO(3)\rightarrow\mathbb{R}^{2}, \\
&\mathbf{q}_{1},\mathbf{q}_{2}\mapsto\mathbf{N}(\mathbf{q}_{2})^{T}\boldsymbol{\theta}(\mathbf{n}(\mathbf{q}_{1}),\mathbf{n}(\mathbf{q}_{2})),
\end{align}
where $\exp(\cdot)$ is the exponential map of $SO(3)$; $\mathbf{q}_{1},\;\mathbf{q}_{2}\in SO(3)$, $\mathbf{a}\in \mathbf{R}^{2}$, and the function $\boldsymbol{\theta}(\cdot,\cdot)$ returns the minimal rotation vector between two unit vectors:
\begin{equation}
\boldsymbol{\theta}(\boldsymbol{\rho}_{1},\boldsymbol{\rho}_{2})=\arccos(\boldsymbol{\rho}_{1}^{T}\boldsymbol{\rho}_{2})\frac{\boldsymbol{\rho}_{1}\times\boldsymbol{\rho}_{2}}{||\boldsymbol{\rho}_{1}\times\boldsymbol{\rho}_{2}||}\in \mathbb{R}^{3}.
\end{equation}


\subsection{Spherical Image Kinematics}

Consider a stationary 3D landmark point $l$. Let $\mathbf{q}$ be the corresponding detected feature in the camera frame and $r$ be the range from the camera origin to point $l$. Our goal is to obtain the time derivatives of $\mathbf{q}$ denoted as $\dot{\mathbf{q}}$ and the time derivatives of $r$ denoted as $\dot{r}$.

From Fig. \ref{fig_frame}, we know $\mathbf{p}_{lw}^{w}=\mathbf{C}(\mathbf{q}_{wc})\mathbf{p}_{lc}^{c}+\mathbf{p}_{cw}^{w}$. Differentiating this equation with respect to time on both sides, inserting $\mathbf{p}_{lc}^{c}=\mathbf{n}(\mathbf{q})r$, and applying the chain rule yield:
\begin{align}
\begin{split}
\mathbf{0}&=\frac{d}{dt}(\mathbf{p}_{lw}^{w})=\frac{d}{dt}(\mathbf{C}(\mathbf{q}_{wc})(\mathbf{n}(\mathbf{q})r)+\mathbf{p}_{cw}^{w})\\
&=\mathbf{\dot{p}}_{cw}^{w}\!\!+\!\!\dot{\mathbf{C}}(\mathbf{q}_{wc})\mathbf{n}(\mathbf{q})r\!\!+\!\!\mathbf{C}(\mathbf{q}_{wc})\frac{\partial\mathbf{n}(\mathbf{q})}{\partial\mathbf{q}}\dot{\mathbf{q}}r\!\!+\!\!\mathbf{C}(\mathbf{q}_{wc})\mathbf{n}(\mathbf{q})\dot{r}.
\end{split} \label{equ_image_dynamics}
\end{align}
Some of the derivatives in (\ref{equ_image_dynamics}) can be obtained via three-dimensional geometry \cite{barfoot2017state}:
\begin{align}
\mathbf{\dot{p}}_{cw}^{w}&=\mathbf{v}_{cw}^{w},\\
\dot{\mathbf{C}}(\mathbf{q}_{wc})&=\boldsymbol{\omega}_{cw}^{w^{\times}}\mathbf{C}(\mathbf{q}_{wc}),
\end{align}
where $\mathbf{v}_{cw}^{w}$ denotes the velocity of the robot expressed in frame $\mathcal{F}^{w}$ and $\boldsymbol{\omega}_{wc}^{w}$ denotes the angular velocity from frame $\mathcal{F}^{w}$ to frame $\mathcal{F}^{c}$, expressed in frame $\mathcal{F}^{w}$. The most tricky one, $\frac{\partial}{\partial\mathbf{q}}\mathbf{n}(\mathbf{q})$, can be derived as follows:
\begin{align}
	\begin{split}\frac{\partial}{\partial\mathbf{q}}\mathbf{n}(\mathbf{q}) & =\underset{\varepsilon\rightarrow0}{\lim}\left[\begin{array}{c}
			(\frac{\mathbf{n}(\mathbf{q}\boxplus(e_{1}\varepsilon))-\mathbf{n}(\mathbf{q})}{\varepsilon})^{T}\\
			(\frac{\mathbf{n}(\mathbf{q}\boxplus(e_{2}\varepsilon))-\mathbf{n}(\mathbf{q})}{\varepsilon})^{T}
		\end{array}\right]^{T}\\
		& \approx\underset{\varepsilon\rightarrow0}{\lim}\left[\begin{array}{c}
			(\frac{(\mathbf{I}+(\mathbf{N}(\mathbf{q})\mathbf{e}_{1}\varepsilon)^{\times})\mathbf{C}(\mathbf{q})\mathbf{e}_{z}-\mathbf{C}(\mathbf{q})\mathbf{e}_{z}}{\varepsilon})^{T}\\
			(\frac{(\mathbf{I}+(\mathbf{N}(\mathbf{q})\mathbf{e}_{2}\varepsilon)^{\times})\mathbf{C}(\mathbf{q})\mathbf{e}_{z}-\mathbf{C}(\mathbf{q})\mathbf{e}_{z}}{\varepsilon})^{T}
		\end{array}\right]^{T}\\
		& =-\mathbf{n}(\mathbf{q})^{\times}\mathbf{N}(\mathbf{q})\in\mathbb{R}^{3\times2},
	\end{split}
\end{align}
where $\varepsilon\in\mathbf{R}$ is the small-angle perturbation and $\mathbf{e}_{1/2}\in\mathbb{R}^{2}$ are orthonormal basis vectors. Here the identity $\mathbf{v}_{1}^{\times}\mathbf{v}_{2}=-\mathbf{v}_{2}^{\times}\mathbf{v}_{1}$ is used where $\mathbf{v}_{1/2} \in \mathbb{R}^{3}$. The small-angle approximation is also used: $\mathbf{C}(\exp(\Delta\boldsymbol{\psi})\otimes\mathbf{q})\approx(\mathbf{I}+\Delta\boldsymbol{\psi}^{\times})\mathbf{C}(\mathbf{q})$ where $\Delta\boldsymbol{\psi}\in\mathbb{R}^{3}$.

We define $\mathbf{v}^{c}=\mathbf{C}(\mathbf{q}_{wc})^{T}\mathbf{v}_{cw}^{w}$ and $\boldsymbol{\omega}^{c}=\mathbf{C}(\mathbf{q}_{wc})^{T}\boldsymbol{\omega}_{cw}^{w}$. Inserting (9-11) and $(\mathbf{v}^{c},\boldsymbol{\omega}^{c})$ into (\ref{equ_image_dynamics}), followed by left multiplying $\mathbf{C}(\mathbf{q}_{wc})^T$ on both sides, we get:
\begin{equation}
	\mathbf{0}=\mathbf{v}^{c}-\mathbf{n}(\mathbf{q})^{\times}\boldsymbol{\omega}^{c}r-\mathbf{n}(\mathbf{q})^{\times}\mathbf{N}(\mathbf{q})\dot{\mathbf{q}}r+\mathbf{n}(\mathbf{q})\dot{r}. \label{equ_image_dynamics_simplified}
\end{equation}
Here the identity $(\mathbf{C}(\mathbf{q}_{wc})\mathbf{n}(\mathbf{q}))^{\times}=\mathbf{C}(\mathbf{q}_{wc})\mathbf{n}(\mathbf{q})^{\times}\mathbf{C}(\mathbf{q}_{wc})^{T}$ is used to simplify the right-hand-side of (\ref{equ_image_dynamics}).

Finally, we obtain the following the image kinematics as well as the range kinematics:
\begin{align}
	\dot{\mathbf{q}}&=\mathbf{N}(\mathbf{q})^{T}(-\mathbf{n}(\mathbf{q})^{\times}\frac{\mathbf{v}^{c}}{r}-\boldsymbol{\omega}^{c}),\label{equ_image_dynamics_final}\\
	\dot{r}&=-\mathbf{n}(\mathbf{q})^{T}\mathbf{v}^{c}. \label{equ_range_dynamics_final}
\end{align}
by pre-multiplying (\ref{equ_image_dynamics_simplified}) with $\frac{\mathbf{N}(\mathbf{q})^{T}\mathbf{n}(\mathbf{q})^{\times}}{r}$ and $\mathbf{n}(\mathbf{q})^{T}$, respectively. Here the identity $\mathbf{N}(\mathbf{q})^{T}\mathbf{n}(\mathbf{q})^{\times}\mathbf{n}(\mathbf{q})^{\times}=-\mathbf{N}(\mathbf{q})^{T}$ is used for element eliminations. Note that there is a slight difference in the image kinematics compared to \cite{bloesch2015robust}. This is because we adopt a different robotics convention which is more commonly used in quadrotor control.

The discrete-time feature prediction can be performed by using $\mathbf{q}_{k+1}=\mathbf{q}_{k}\boxplus\left(\mathbf{N}(\mathbf{q})^{T}(-\mathbf{n}(\mathbf{q})^{\times}\frac{\mathbf{v}^{c}}{r}-\boldsymbol{\omega}^{c})\Delta t\right)$, which can also be extended as
\begin{equation}
\mathbf{q}_{k+1}=\exp((-\mathbf{n}(\mathbf{q})^{\times}\frac{\mathbf{v}^{c}}{r}\!-\!(\mathbf{I}-\mathbf{n}(\mathbf{q})\mathbf{n}(\mathbf{q})^{T})\boldsymbol{\omega}^{c})\Delta t)\otimes\mathbf{q}_{k},
\end{equation}
where $k$ is the time step and $\Delta t$ is the time gap. Here the identity $\mathbf{N}(\mathbf{q})\mathbf{N}(\mathbf{q})^{T}=(\mathbf{I}-\mathbf{n}(\mathbf{q})\mathbf{n}(\mathbf{q})^{T})$ is used.


\section{Model Predictive Visual Servoing}
\begin{figure}[!t]
	\centering
	\includegraphics[width=0.45\textwidth]{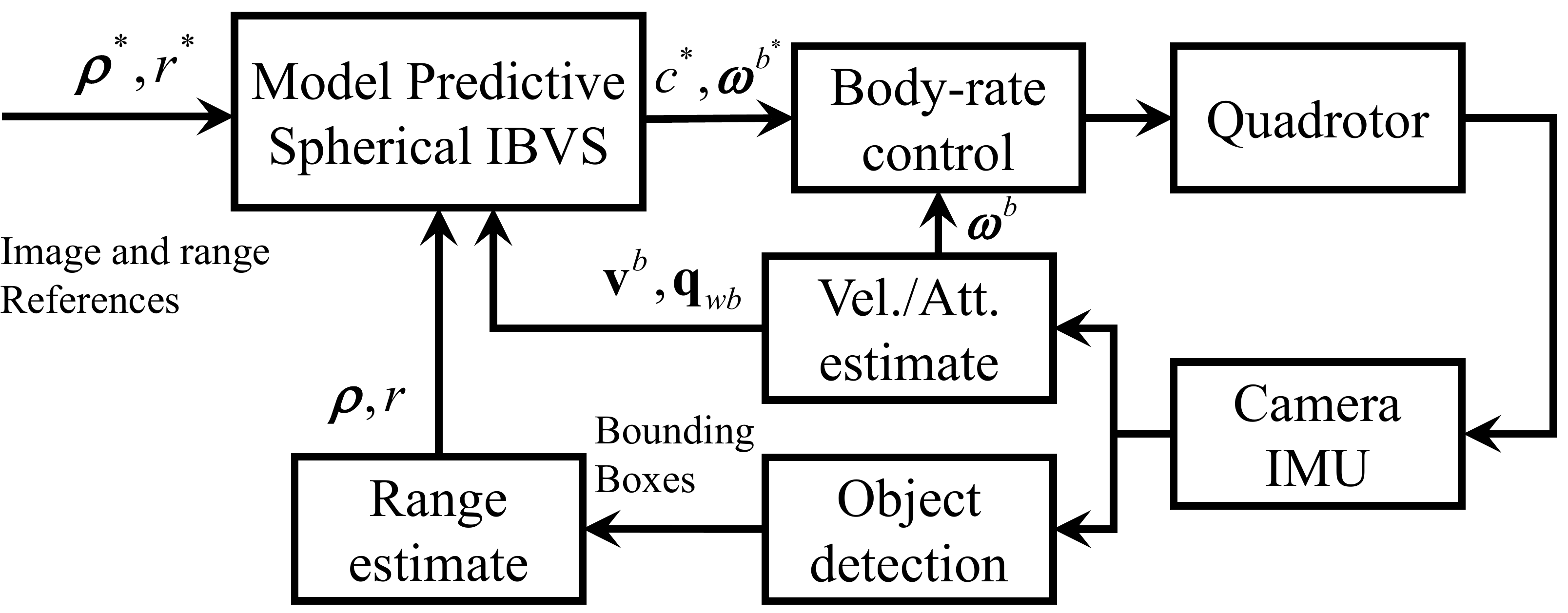}
	\caption{Control diagram of the the proposed IBVS algorithm.}
	\label{fig_diagram}
\end{figure}

This section introduces the NMPC framework for our spherical IBVS. The overall system is illustrated in Fig. \ref{fig_diagram}.

\subsection{Optimal Control Problem}
The non-linear optimization problem can be formulated as:
\begin{align}
	\begin{split}
		\min_{\mathbf{x},\mathbf{u},\mathbf{z}}\quad & \int_{t_{0}}^{t_{f}}(\mathcal{L}_{a}(\mathbf{x},\mathbf{u})+\mathcal{L}_{p}(\mathbf{x},\mathbf{u})+\mathcal{L}_{z}(\mathbf{z}))dt\\
		\textrm{s.t.} \quad
		& \dot{\mathbf{x}}=\mathbf{f}(\mathbf{x},\mathbf{u}),\\
		&\mathbf{h}(\mathbf{x},\mathbf{u},\mathbf{z})\leqslant\mathbf{0},
	\end{split}\label{equ_mpc}
\end{align}
where $t_0$ is the start time and $t_f$ is the end time. $\mathbf{x}$, $\mathbf{u}$, and $\mathbf{z}$ represent state variables, control inputs, and slack variables, respectively. The state dynamics $\dot{\mathbf{x}} = \mathbf{f}(\mathbf{x}, \mathbf{u})$ consists of the image kinematics (\ref{equ_image_dynamics_final}), range kinematics (\ref{equ_range_dynamics_final}), and quadrotor dynamics introduced below. The inequality constraint $\mathbf{h}(\mathbf{x},\mathbf{u},\mathbf{z})\leqslant\mathbf{0}$ includes the robot's physical constraint, target-visibility constraint, and TTC constraint. Three objective functions are designed to fulfill the flight mission: the action objectives $\mathcal{L}_{a}(\mathbf{x},\mathbf{u})$, the perception objectives $\mathcal{L}_{p}(\mathbf{x},\mathbf{u})$, and the slack penalties $\mathcal{L}_{z}(\mathbf{z})$. 

\subsection{State Dynamics}
The state and control input vectors are defined as:
\begin{align}
	\mathbf{x} & =[\mathbf{v}^{w},\mathbf{q}_{wb},\mathbf{q},r]^{T},\label{equ_state}\\
	\mathbf{u} & =[c,\boldsymbol{\omega}^{b}]^{T}\label{equ_control},
\end{align}
where $\mathbf{v}^{w}$ is the simplified notation of $\mathbf{v}_{cw}^{w}$; $\boldsymbol{\omega}^{b}$ is the simplified notation of $\boldsymbol{\omega}_{bw}^{b}$, a.k.a. the body rates of the quadrotor; $\mathbf{q}_{wb}$ denotes the quadrotor orientation; and $c$ denotes the mass-normalized thrust. We assume that all the state variables in (\ref{equ_state}) can be measured or estimated from the input image.

The full state dynamics can be constructed as:

\begin{align}
\begin{split}
	\mathbf{\dot{v}}^{w}&=\mathbf{C}(\mathbf{q}_{wb})\mathbf{c}+\mathbf{g}^{w},\\
	\dot{\mathbf{q}}_{wb}&=\frac{1}{2}\left[\begin{array}{c}
		0\\
		\boldsymbol{\omega}^{b}
	\end{array}\right]\otimes\mathbf{q}_{wb},\label{equ_quadrotor}
\end{split}\\
\begin{split}
	\dot{\mathbf{q}}&=\mathbf{N}(\mathbf{q})^{T}(-\frac{1}{r}\mathbf{n}(\mathbf{q})^{\times}\mathbf{C}(\mathbf{q}_{bc})^{T}(\mathbf{C}(\mathbf{q}_{wb})^{T}\mathbf{v}^{w}+\\&\quad \quad\boldsymbol{\omega}^{b^{\times}}\mathbf{p}_{cb}^{b}) -\mathbf{C}(\mathbf{q}_{bc})^{T}\boldsymbol{\omega}^{b}),\\
	\dot{r}&=-\mathbf{n}(\mathbf{q})^{T}\mathbf{C}(\mathbf{q}_{bc})^{T}(\mathbf{C}(\mathbf{q}_{wb})^{T}\mathbf{v}^{w}+\boldsymbol{\omega}^{b^{\times}}\mathbf{p}_{cb}^{b})\label{equ_image}
\end{split}
\end{align}
where (\ref{equ_quadrotor}) corresponds to the quadrotor dynamics \cite{falanga2018pampc} and (\ref{equ_image})  corresponds to the image and range kinematics after mapping $\mathbf{v}^{c}$ to frame $\mathcal{F}^{w}$ and $\boldsymbol{\omega}^{c}$ to frame $\mathcal{F}^{b}$, respectively; $\mathbf{g}^{w}=(0,0,-9.81)$ denotes the gravity vector expressed in frame $\mathcal{F}^{w}$; and $\mathbf{c}=(0,0,c)^{T}$.

\subsection{Objective Functions}

%
%
%
%

To keep a specified position relative to the target, we need to retrieve positional information from the current target-feature measurement $\mathbf{q}_0$ as well as its future location $\mathbf{q}_k$ predicted at time step $k=1,...,N$ where $N$ is the horizon of NMPC. This can be done by compensating for the time-varying quadrotor rotation $\mathbf{q}_{wb_{k}}$ predicted at the corresponding time step. Let $\mathbf{p}^{*}$, $\mathbf{v}^{*}$, $\mathbf{q}_{wb}^{*}$, and $\mathbf{u}^{*}$ be the reference relative-position, velocity, orientation, and control inputs, respectively. We define the following action objective function to fulfill quadrotor stabilization, rotation compensation, and target tracking:
\begin{equation}
\begin{split}
	\mathcal{L}_{a}(\mathbf{x},\mathbf{u})=\Vert\mathbf{p}^{*}-\mathbf{p}_{b_{k}l}^{w}\Vert_{\mathbf{Q}_{p}}^{2}+\Vert\mathbf{v}^{*}-\mathbf{v}_{k}^{w}\Vert_{\mathbf{Q}_{v}}^{2}+\\
	\Vert\mathbf{q}_{wb}^{*}-\mathbf{q}_{wb_{k}}\Vert_{\mathbf{Q}_{q}}^{2}+\Vert\mathbf{u}^{*}-\mathbf{u}_{k}\Vert_{\mathbf{R}}^{2},
\end{split}
\end{equation}
where
\begin{equation}
\mathbf{p}_{b_{k}l}^{w}=\mathbf{C}(\mathbf{q}_{wb_{k}})(\mathbf{C}(\mathbf{q}_{bc})\mathbf{n}(\mathbf{q}_{k})r_{k}+\mathbf{p}_{cb}^{b}).
\end{equation}
What we do here can be summarized as predicting the feature motion using the state kinematics, aligning the predicted features to the inertial frame $\mathcal{F}^{w}$, and then formulating position error $\Vert\mathbf{p}^{*}-\mathbf{p}_{b_{k}l}^{w}\Vert_{\mathbf{Q}_{p}}^{2}$ like PBVS to achieve global convergence. The reason we prefer a position error instead of an image error with a range error, e.g., $\Vert\boldsymbol{\rho}^{*}-\mathbf{C}(\mathbf{q}_{wb_{k}}\otimes\mathbf{q}_{bc})\mathbf{n}(\mathbf{q}_{k})\Vert_{\mathbf{Q}_{\rho}}^{2}+\Vert r^{*}-r_{k}\Vert_{\mathbf{Q}_{r}}^{2}$, is that we want to account for the unit mismatch between the image error and range error; by scaling the image error with a range prediction, we obtain a 3D position metric with each element sharing the same unit. This can effectively facilitate the tuning of the optimization weights. Moreover, by replacing a fixed $\mathbf{p}^{*}$ with a time-varying $\mathbf{p}^{*}(t)$, we can benefit from a path-planning algorithm, and a larger region of convergence can be gained.

We define the following perception objective to force the camera to face the target:
\begin{equation}
	\mathcal{L}_{p}(\mathbf{x},\mathbf{u})=\Vert\mathbf{n}(\mathbf{q}_{k})_{x}/\mathbf{n}(\mathbf{q}_{k})_{z}-0\Vert_{\mathbf{Q}_{u}}^{2},\label{euq_perception_obj}
\end{equation}
where $\mathbf{n}(\mathbf{q}_{k})_{x/y/z}$ corresponds to the element in the x/y/z-axis. Here we first project the feature from the spherical coordinate to the homogeneous coordinate and then 
define an error in the horizontal image axis between the current feature and the image center. Note that we do not add a similar objective function in the vertical image axis as in \cite{falanga2018pampc} because it will result in a conflict between the action objective and the perception objective. Lastly, readers can refer to \cite{verschueren2022acados} for detailed formulation of the slack penalty $\mathcal{L}_{z}(\mathbf{z})$.

\subsection{Visibility Constraint}

To keep the feature inside the specified image bound, we model the visibility constraint as
\begin{align}
	\begin{split}
		-\mathbf{s}_{max}-z&\leqslant\mathbf{n}(\mathbf{q}_{k})_{x}/\mathbf{n}(\mathbf{q}_{k})_{z}\leqslant\mathbf{s}_{max}+z,\\
		-\mathbf{s}_{max}-z&\leqslant\mathbf{n}(\mathbf{q}_{k})_{y}/\mathbf{n}(\mathbf{q}_{k})_{z}\leqslant\mathbf{s}_{max}+z,
	\end{split}
\end{align}
where $z$ denotes the slack variable. Since this is a soft constraint and it is still possible for the feature to exceed the bound, $\mathbf{s}_{max}>0$ should be smaller than the actual image size to maintain a safe margin to the image edge. The main reason for not using a hard constraint is to reduce sensitivity to noise. Our practices show that it is very easy for a hard visibility constraint to cause an optimization failure when the target is close and the camera rotation is large. This is because a small rotation can result in a large feature motion if $r$ is small according to (\ref{equ_image_dynamics_final}) and any noise may lead to an infeasible solution.


%
%
%

\subsection{Time-to-Collision Constraint}
Humans can perceive an informational variable called time to collision---the required time to collide with an object if the current velocity is maintained---to avoid potential collision \cite{zhang2018optimal}. TTC can be expressed as:
\begin{equation}
	t_{c}=r/\dot{r}.
\end{equation}

We model TTC as a soft constraint to prevent the drone to be too close to the target. Let $\mathcal{T}_{c}=\{t_{c}\in\mathbb{R}|t_{c}<0\;\text{or}\;t_{c}\geqslant t_{c_{min}}\}$ be the safe $t_{c}$ set where $t_{c_{min}}$ is the minimum TTC. It can be interpreted as the flight is safe if the robot is leaving the object ($t_{c}<0$) or it takes a time longer than $t_{c_{min}}$ to hit the object. To keep $t_{c}$ inside the feasible set $\mathcal{T}_{c}$, we define the TTC constraint as:
\begin{equation}
	\dot{r}/r\leqslant 1/t_{c_{min}}+z.
\end{equation}
Compared with the distance (range) constraint as in \cite{ji2022elastic}, the proposed TTC constraint can provide an additional speed regulation which is very effective in reducing tracking overshoot in a high-speed flight.

\subsection{Implementation Details}
The system is implemented in C++ and runs in ROS environment. The NMPC framework is based on ACADO with qpOASES as the solver \cite{houska2011acado}. We set the time gap as $\Delta t = 0.05$ s with a horizon of $N = 20$. To account for the drifts in feature prediction, our framework works in a receding-horizon fashion by iteratively solving the optimization problem. In order to deal with the intermittent object-detection outage, we store all the control inputs after each iteration and broadcast them in order if the new detection result is not received on time.

%

\section{Simulation}
We simulate a hummingbird quadrotor using RotorS \cite{furrer2016rotors}. A proportional-integral-derivative (PID) controller is developed for low-level body rate tracking. The target is simulated as a movable 3D point and a pinhole camera model with an image size of $752 \times 480 \;\text{px}^2$ is used to generate image feedback.


\begin{figure}[!htbp]
	\centering
	\subfigure[]{\includegraphics[width=0.225\textwidth]{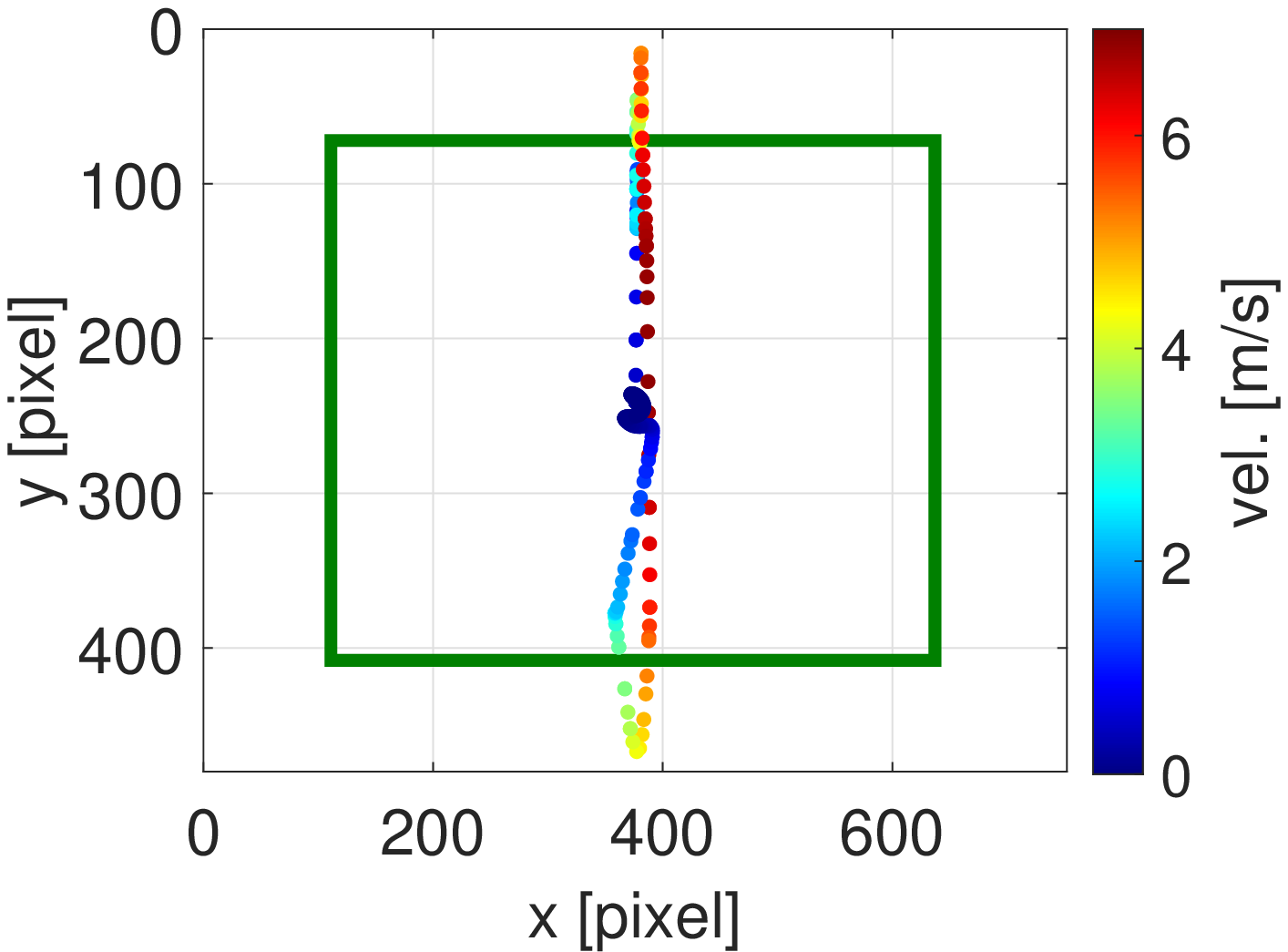}\label{fig_static_feature}}
	\hfil
	\subfigure[]{\includegraphics[width=0.225\textwidth]{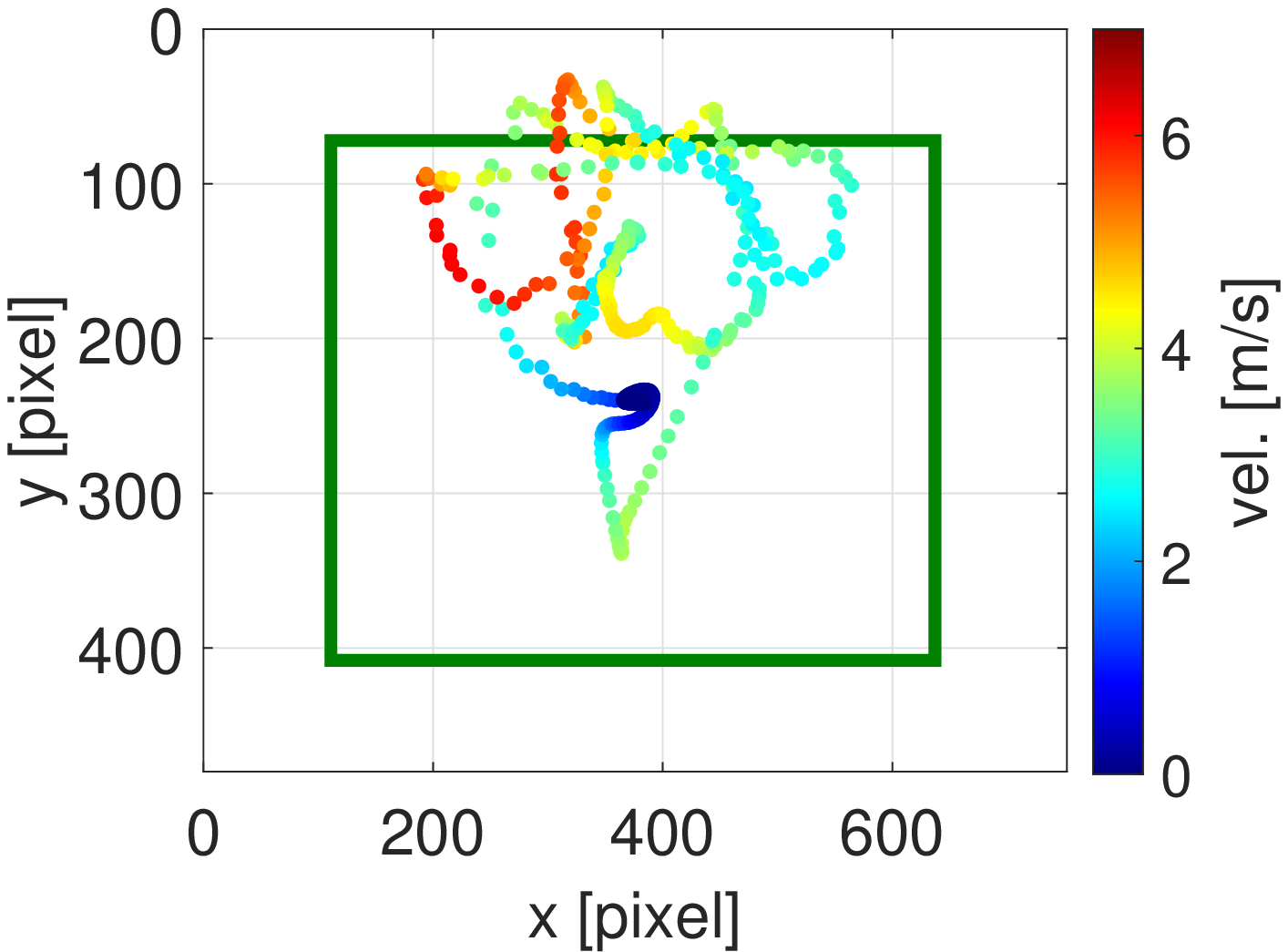}\label{fig_moving_feature}}
	\caption{Observed features in the image plane. The specified image bound is visualized as a green rectangle. The color of the feature point indicate the speed of the robot when the feature is being captured. (a) Features observed in the stationary target tracking scenario. (b) Features observed in the moving target tracking scenario.}
	\label{fig_feature}
\end{figure}

\subsection{Stationary Target Tracking}\label{sec_stationary_target}
In this scenario, the quadrotor is commanded to reach a stationary target from 20 m away. The target's position is $(20,0,1)$ m and the quadrotor's initial position is $(0,0,1)$ m with velocity and attitude both being zeros. The desired range is set to be $r^{*}=2$ m and the reference feature is the image center $\boldsymbol{\rho}^{*}=(0,0,1)^T$. The image bound is specified as a rectangle centered at the middle of the image with a size of $557 \times 336 \;\text{px}^2$. The minimum TTC is set to be $t_{c_{min}}=2$ s. 

\begin{figure}[!htbp]
	\centering\
	\includegraphics[width=0.35\textwidth]{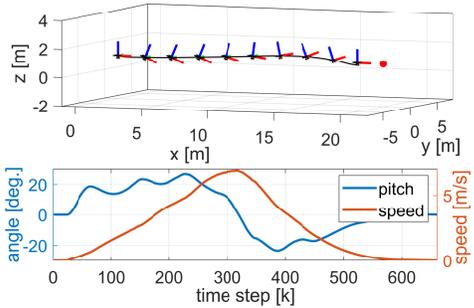}
	\caption{Quadrotor trajectory in the stationary target tracking scenario with the target visualized as a red ball. The body frame is indicated by \{\textcolor{red}{$\mathbf{x}$}, \textcolor{green}{$\mathbf{y}$}, \textcolor{blue}{$\mathbf{z}$}\}. The time evolution of the pitch angle and the speed of the quadrotor is plot in the bottom figure.}
	\label{fig_static_traj}
\end{figure}

Fig. \ref{fig_static_feature} shows the observed feature in the image plane during the whole task and Fig. \ref{fig_static_traj} provides the tracking trajectory of the quadrotor. As expected, the robot had successfully reached the reference state without any feature leaving the FOV. In addition, we do observe that some features are outside the image bound (green box). This is because the visibility constraint is a soft constraint that can be violated temporally if the controller prioritizes aggressiveness in certain stages such as an acceleration stage at the beginning and a deceleration stage at the end. We can also let the controller strictly obey the visibility constraint by setting a large weight for the slack variable. According to Fig. \ref{fig_static_traj}, we know that the maximum pitch angle of the quadrotor reaches 26.78$^{\circ}$ with a maximum speed of 6.88 m/s. This result suffices to show that our method can produce aggressive quadrotor flights and can be applicable for high-speed tasks.

%


\subsection{Moving Target Tracking}

\begin{figure}[!t]
	\centering
	\subfigure[]{\includegraphics[width=0.35\textwidth]{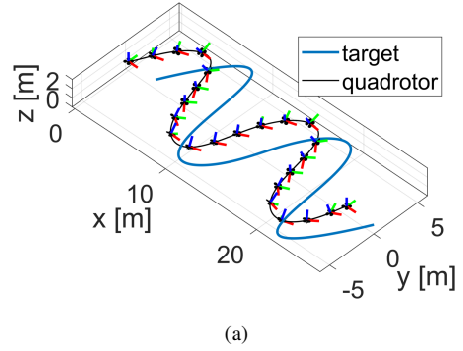}\label{fig_moving_3Dtraj}}
	\hfil
	\subfigure[]{\includegraphics[width=0.3\textwidth]{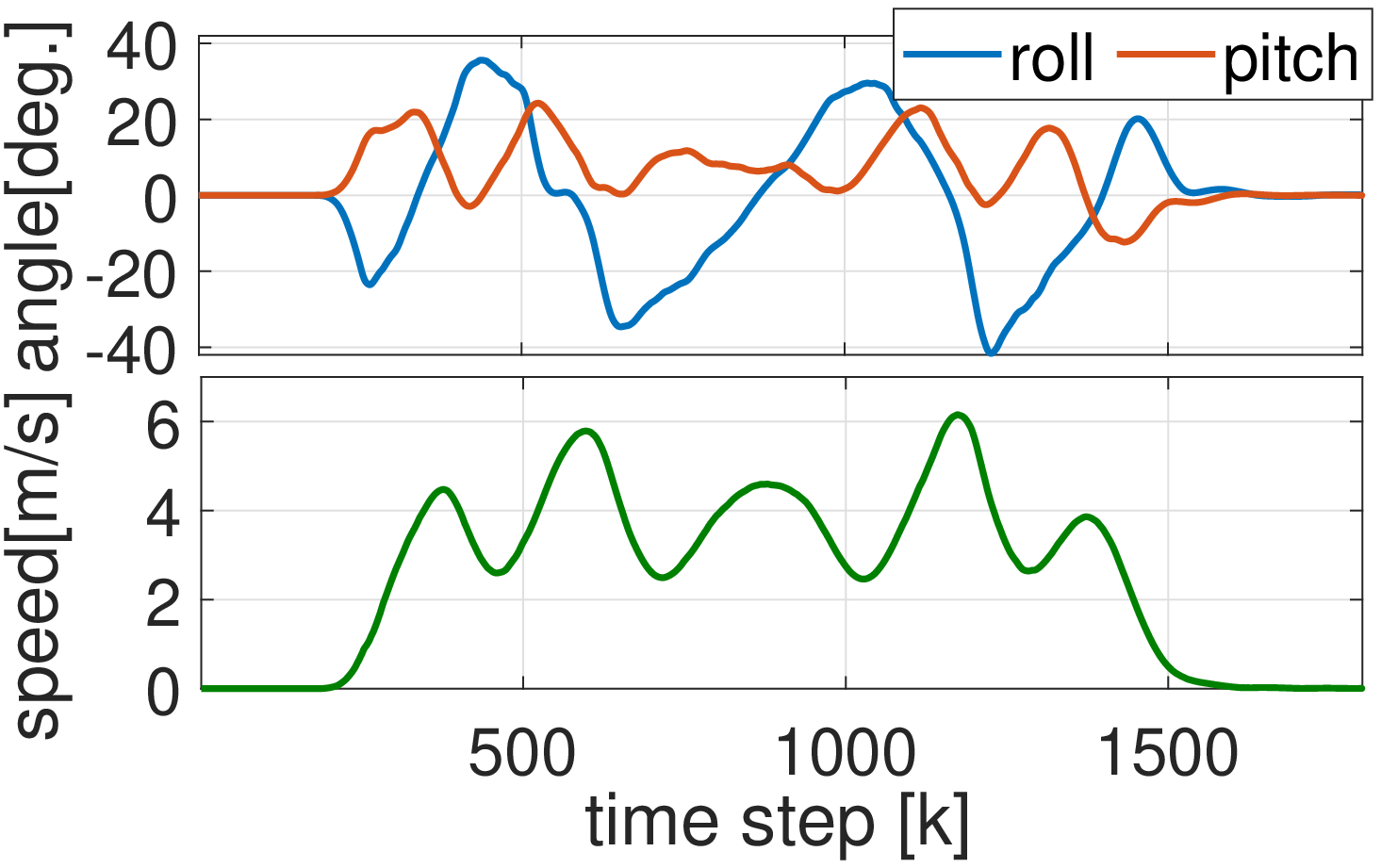}\label{fig_moving_rps}}
	\caption{(a) 3D trajectory of the quadrotor (black line) in the moving target tracking scenario with the blue line representing the target trajectory. The body frame is indicated by \{\textcolor{red}{$\mathbf{x}$}, \textcolor{green}{$\mathbf{y}$}, \textcolor{blue}{$\mathbf{z}$}\}. (b) The time evolution of the roll, pitch, and speed of the quadrotor.}
	\label{fig_moving}
\end{figure}

In this scenario, we let the robot track a target that moves along an "S" shape trajectory. The maximum speed of the target is 6 m/s. The desired range is set to be $r^{*}=3$ m and the reference image feature is the image center $\boldsymbol{\rho}^{*}=(0,0,1)^T$. In the beginning, the range error and image error are both zeros.

The trajectories of the quadrotor and the target are shown in Fig. \ref{fig_moving_3Dtraj}. It shows that the robot had successfully followed the target. From Fig. \ref{fig_moving_rps}, we know that the maximum roll angle is 41.62$^{\circ}$, the maximum pitch angle is 24.26$^{\circ}$, and the maximum tracking speed is 6.15 m/s. This result confirms that the resulting flight trajectory is aggressive enough. From Fig. \ref{fig_moving_feature}, we see that under such a large rotation and a high speed, the target feature can still be visible at all times, showing that the produced aggressive maneuvers would not jeopardize the target visibility and the visibility constraint can take effect in large rotations. Another interesting finding in Fig. \ref{fig_moving_feature} is that there are a lot of features lying around the image's upper bound (green line at the top). It shows that the agility of the quadrotor is mostly restricted by the visibility constraint.

\subsection{TTC Constraint vs. Distance Constraint}

\begin{figure}[!htbp]
	\centering
	\subfigure[]{\includegraphics[width=0.225\textwidth]{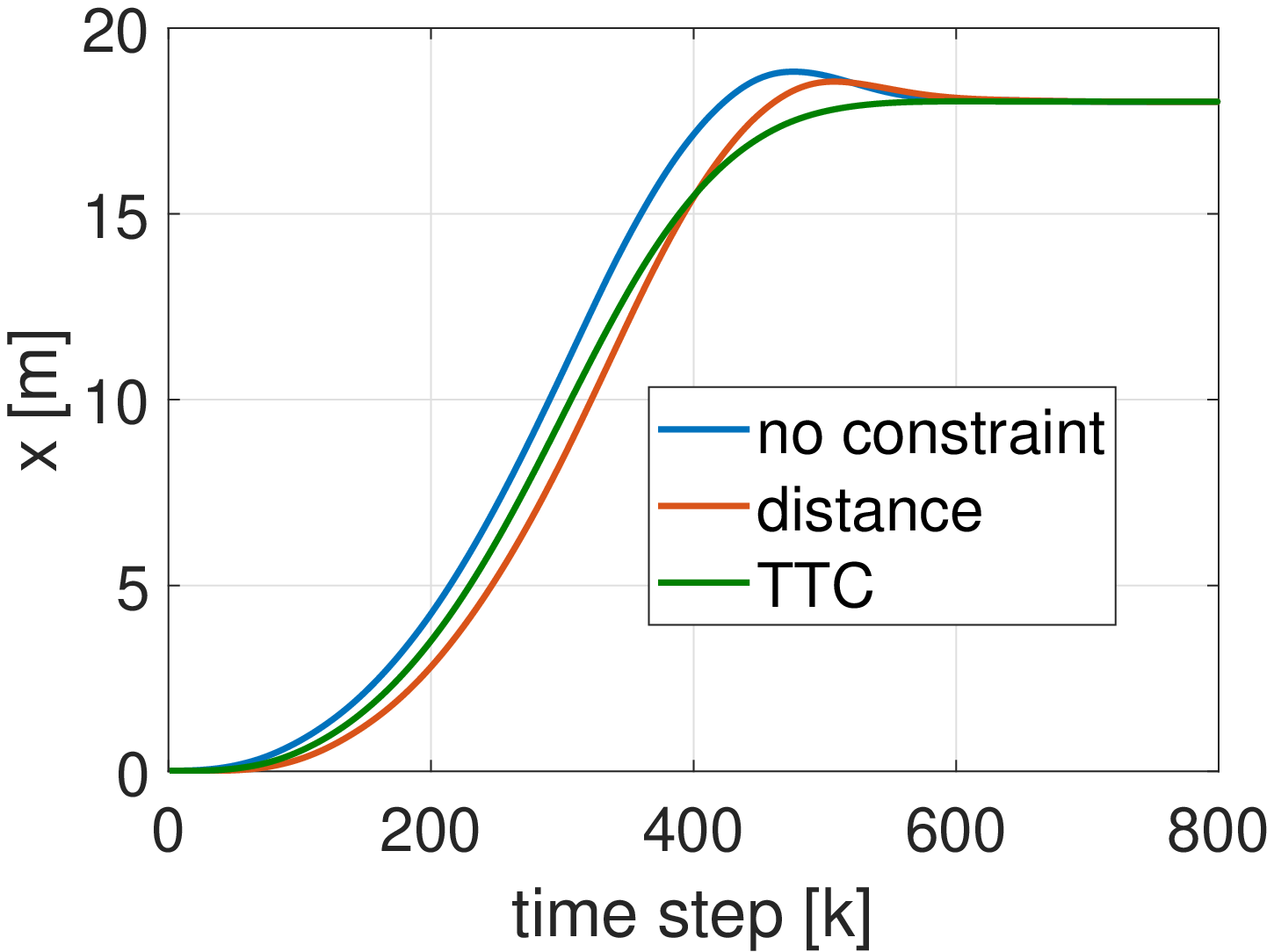}\label{fig_static_overshoot}}
	\hfil
	\subfigure[]{\includegraphics[width=0.225\textwidth]{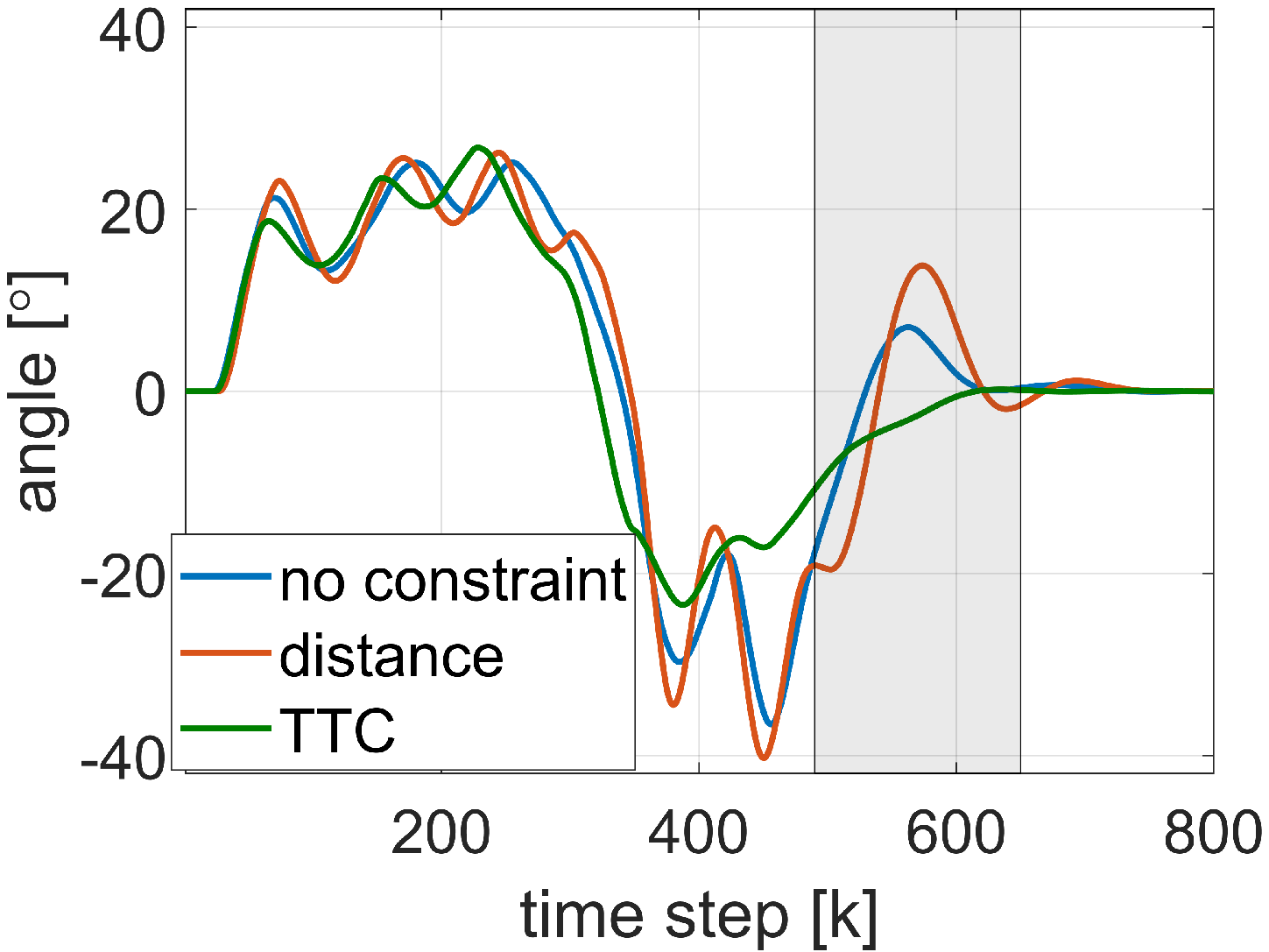}\label{fig_static_ovreshoot_rollpitch}}
	\caption{Comparison of the tracking performance with different constraint strategies. (a) The time evolution of the position in x-axis. (b) The time evolution of the pitch angle with the terminal stage highlighted in grey.}
\end{figure}

To verify the efficacy of the TTC constraint in reducing overshoot during a high-speed flight, we test the controller in the stationary target tracking scenario as in \ref{sec_stationary_target} with and without the TTC constraint and make a comparison with the distance (range) constraint $r\geqslant0$. Other conditions are the same except for the type of constraint. We focus on their respective control performances in terms of overshoot and pitch angles around the terminal stage (when the quadrotor is sufficiently close to the target). 

The time evolution of the quadrotor's position in the x-axis and pitch angles are shown in Fig. \ref{fig_static_overshoot} and Fig. \ref{fig_static_ovreshoot_rollpitch}, respectively. The maximum speeds for all tests are larger than 6.0 m/s. We see that large overshoots occur when either the distance constraint or no constraint is applied, whereas there is almost no overshoot when the TTC constraint is applied. More specifically, the overshoot for the TTC-constraint case is 0.02 m, which is much lower than 0.82 m of the no-constraint case and 0.55 m of the distance-constraint case. The results suggest that the TTC constraint is effective to decrease the tracking overshoot. Moreover, we see from Fig. \ref{fig_static_ovreshoot_rollpitch} that the pitch angle is varying intensively around the terminal stage if the distance constraint is used, which will cause large camera shakes and hinder the target visibility. In comparison, the TTC constraint contributes to a smooth pitch-angle trajectory such that the feature can move in the image plane slowly and smoothly.

%

\section{Experiment}

\begin{figure}[!t]
	\centering\
	\includegraphics[width=0.4\textwidth]{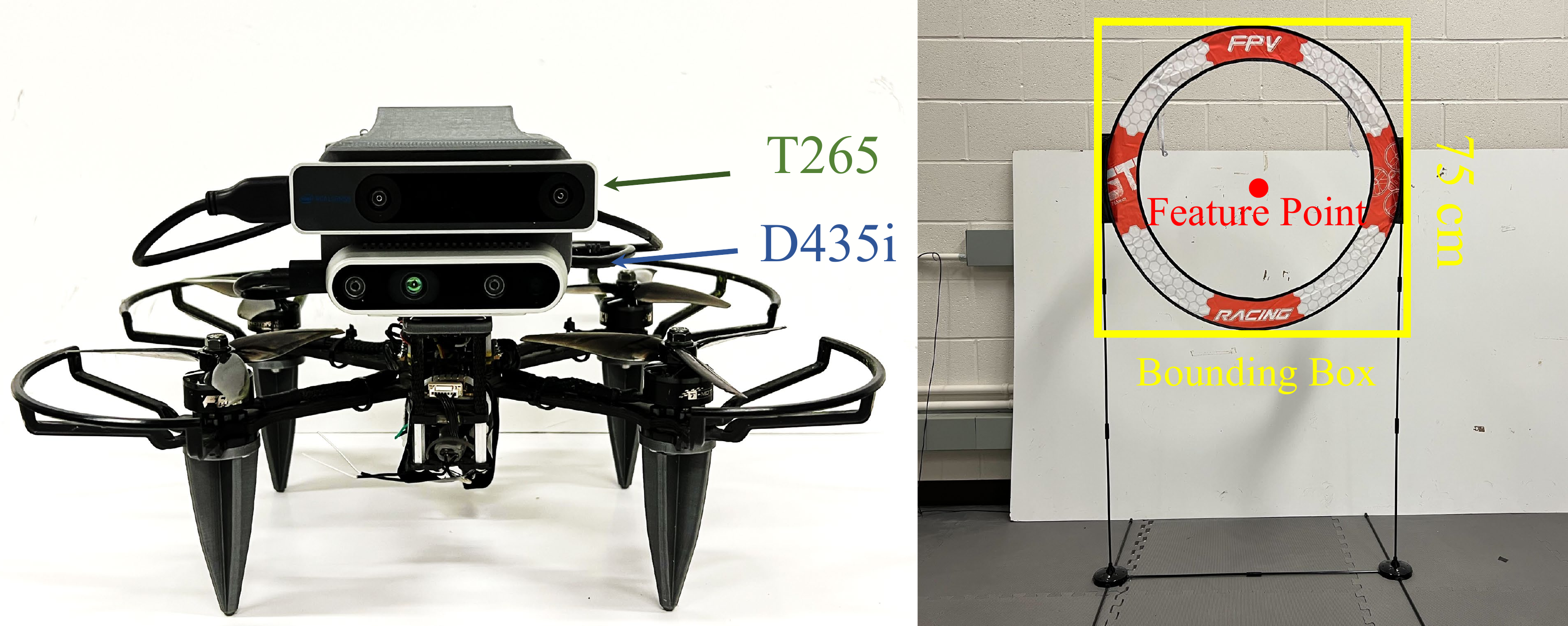}
	\caption{Hardware for real-world experiment. The left figure is our quadrotor platform. The right figure is the target object, a racing gate. The gate can be detected by a deep-learning-based detector which outputs a bounding box as shown in the yellow rectangle. The center point of the bounding box will serve as the feature point.}
	\label{fig_drone}
\end{figure}

\subsection{Hardware}

The quadrotor platform, shown in Fig. \ref{fig_drone}, is equipped with an Intel RealSense D435i depth camera (only the left camera is used) with an image size of $1280 \times 640 \;\text{px}^2$ and an IntelRealsense T265 tracking camera to provide velocity estimate. The onboard computer is a Jetson Xavier NX with its CPU running the control system and its GPU running a deep-learning-based object detector, Yolov5\footnote{https://github.com/ultralytics/yolov5}. The target object is a circular racing gate with a diameter of $d_{gate} = 75$ cm. The detector can output a bounding box of the gate as shown in the yellow box in Fig. \ref{fig_drone}. Using $d_{gate}$ and the size of the detected bounding box, $(w_{box},h_{box})$, we can estimate the range to the gate by using $r=||\boldsymbol{\rho}\cdot d_{gate}/\max(\frac{w_{box}}{f_{cam}},\frac{h_{box}}{f_{cam}})||_{2}$ where $\boldsymbol{\rho}$ is the center of the bounding-box and $f_{cam}$ is the camera's focal length. The low-level body rate controller is based on the Pixhawk4 autopilot. 
\subsection{Real-World Moving Gate Tracking}

\begin{figure}[!htbp]
	\centering
	\subfigure[]{\includegraphics[width=0.225\textwidth]{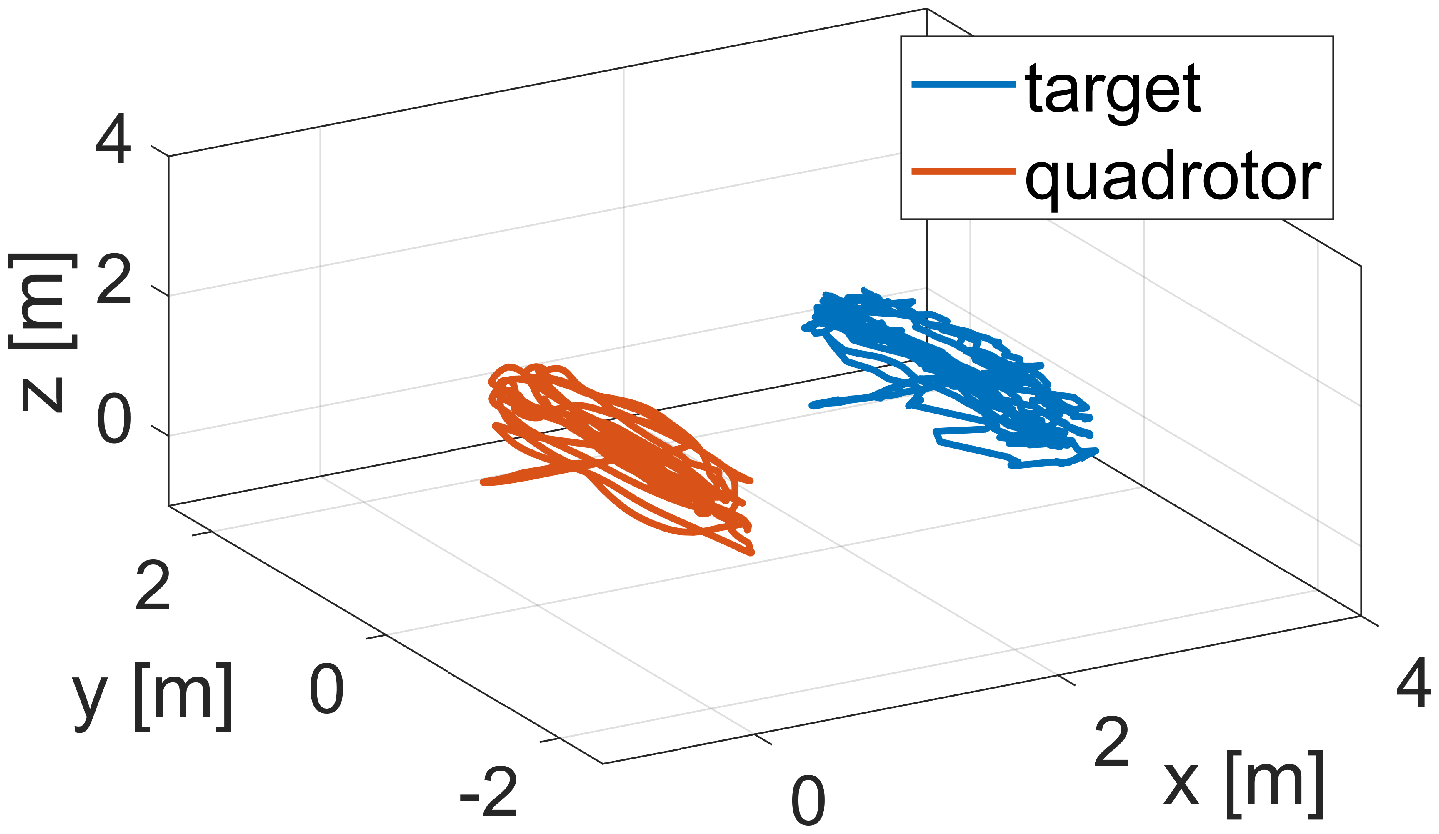}\label{fig_exp_3Dtraj}}
	\hfil
	\subfigure[]{\includegraphics[width=0.225\textwidth]{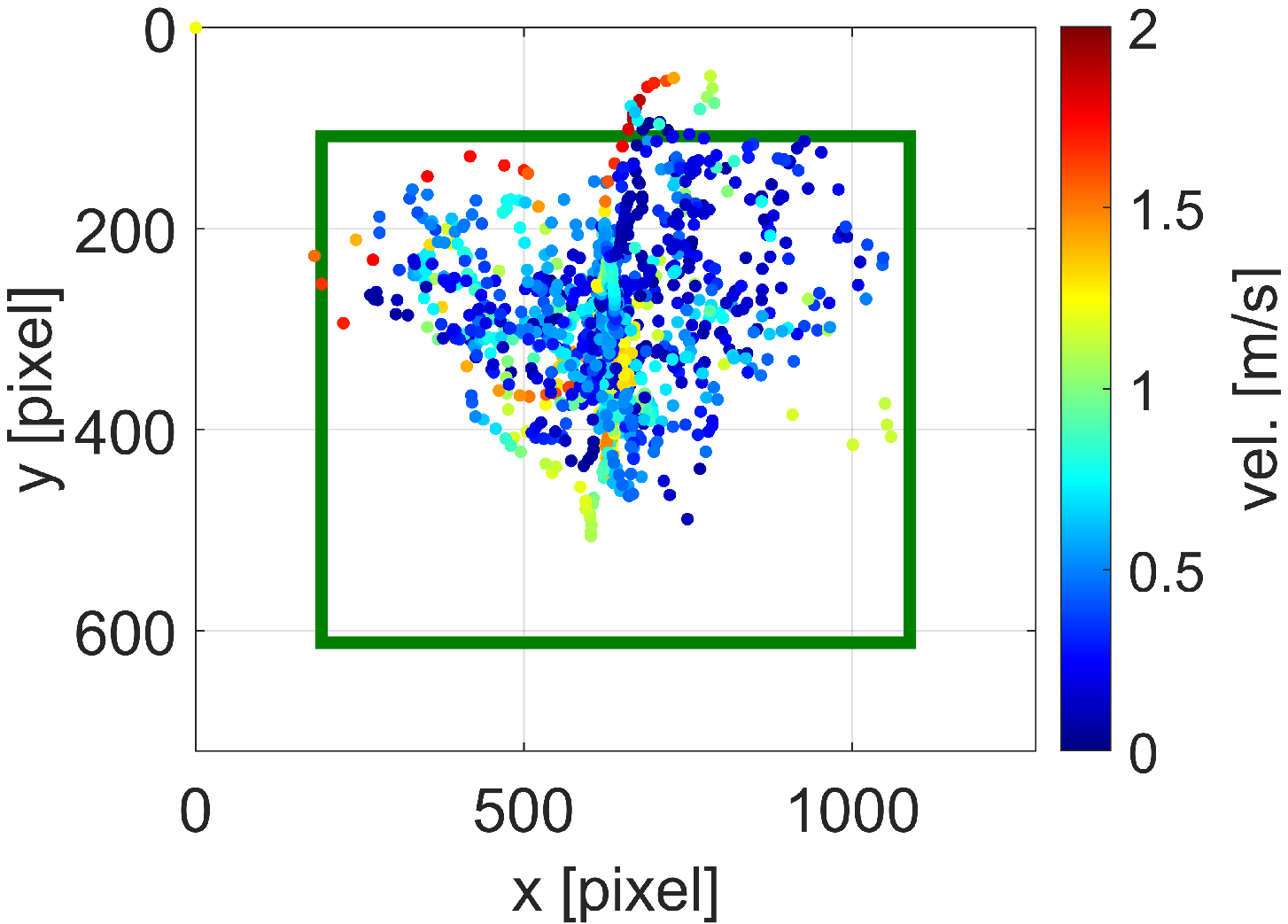}\label{fig_exp_features}}
	\subfigure[]{\includegraphics[width=0.3\textwidth]{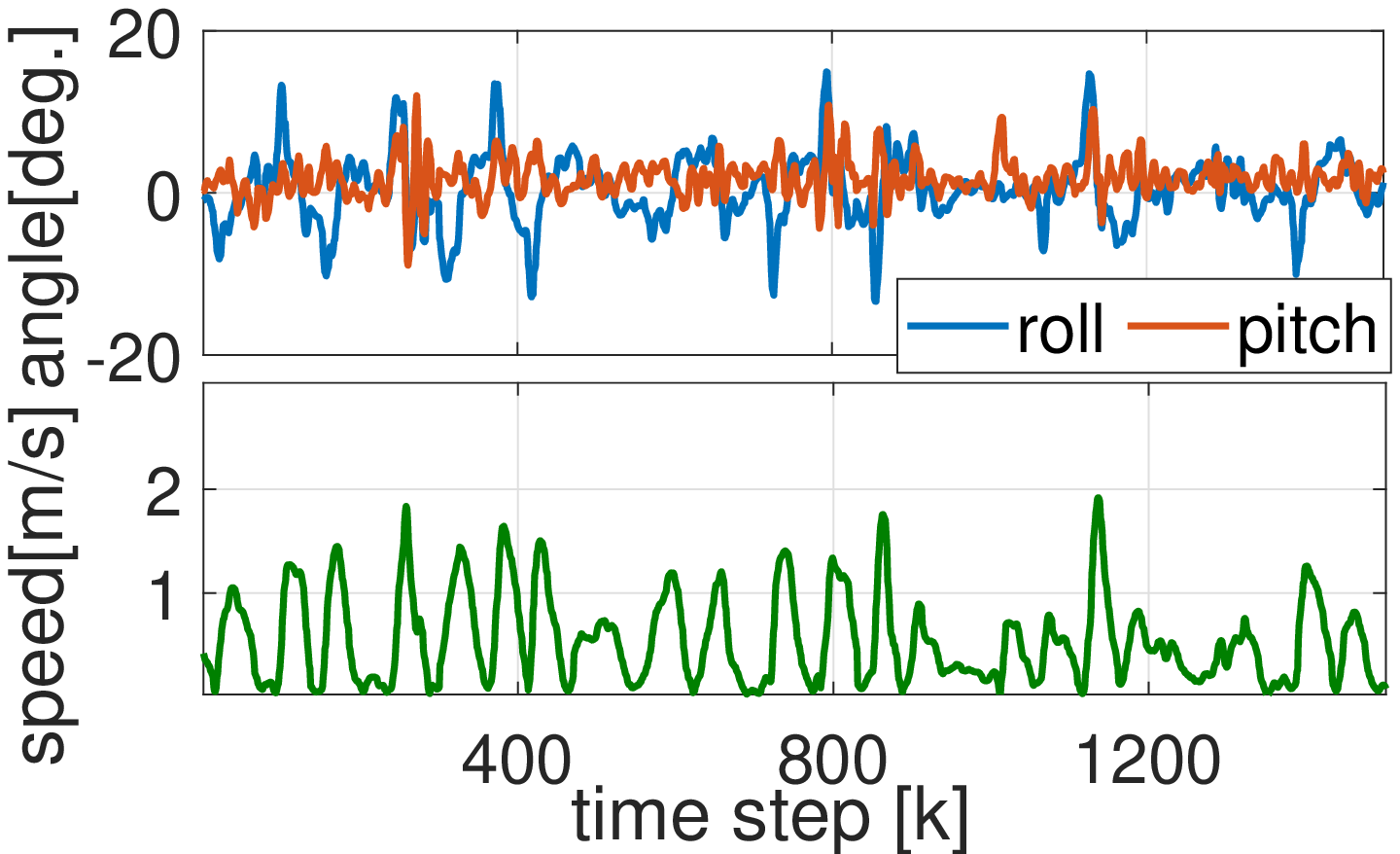}\label{fig_exp_rps}}
	\caption{Experiment results of the moving-gate tracking task. (a) Trajectories of the quadrotor and the target. (b) Observed features in the image plane. The specified image bound is visualized as a green rectangle. The color of the feature point indicate the speed of the robot when the feature is being captured. (c) The time evolution of the roll, pitch, and speed of the quadrotor.}
\end{figure}

This experiment aims to show that our approach can accomplish target tracking stably with pure onboard sensing and computing. The environment is a $5 \;\text{m}\times 4.3 \;\text{m} \times 3 \; \text{m}$ indoor space as shown in Fig. \ref{fig_moving_gate}. The desired range is set to be $r^{*}=2$ m and the reference feature is the image center $\boldsymbol{\rho}^{*}=(0,0,1)^T$. Once the experiment starts, we will move the gate back and forth with a speed up to $2.0$ m/s and evaluate the tracking performance.

The trajectory of the quadrotor and the observed feature in the image plane are displayed in Fig. \ref{fig_exp_3Dtraj} and Fig. \ref{fig_exp_features}, respectively. It shows that the tracking was stable and the target gate can be kept inside the camera FOV even when it was moving rapidly. Fig. \ref{fig_exp_rps} plots the roll and pitch angles as well as the speed of the quadrotor, from which we know that the maximum roll angle is 14.93$^{\circ}$, the maximum pitch-angle is 11.99$^{\circ}$, and the maximum speed is 1.92 m/s. Although the resulting trajectory is not as aggressive as the simulation (mostly limited by the size of the indoor area), the achieved speed is still higher than the existing IBVS methods for a quadrotor \cite{zhang2021robust, mcfadyen2013aircraft}. More importantly, we verify that our framework can empower IBVS to conduct safe and aggressive flights of a quadrotor, which, to some extent, breaks the stereotype that IBVS can only fly a drone in a low-speed and near-hover state.

%
%
%
%



\section{Conclusion}
In this paper, we propose a model predictive spherical IBVS method to conduct aerial tracking with a quadrotor equipped with a front-looking camera and an IMU. We point out the problem of formulating the visibility constraint in the virtual image plane and propose a complete framework to enable modeling the visibility constraint in the actual image plane. Then, we introduce a rotation as an underlying representation of a feature in the spherical coordinate to account for the non-smooth vector field issues as well as the singularity issue. An optimal control problem is formulated in which the image kinematics, range kinematics, and quadrotor dynamics are taken into account. Additionally, we verify the efficacy of the TTC constraint in reducing the tracking overshoot. Extensive simulations and real-world experiments validate the robustness of the proposed controller in moving-target tracking tasks. In the future, we will incorporate collision avoidance to realize image-based aerial tracking in an obstacle-rich environment.
 





\balance
\bibliographystyle{bibtex/bst/IEEEtran}
\bibliography{bibtex/bib/IEEEabrv,bibtex/bib/ibvs}

\addtolength{\textheight}{-12cm}   

\end{document}